\documentclass[conference]{IEEEtran}
\IEEEoverridecommandlockouts
\usepackage{cite}
\usepackage{amsmath,amssymb,amsfonts}
\usepackage{textcomp}
\usepackage{xcolor}
\usepackage{tabularx}
\usepackage[pdftex]{graphicx}
\usepackage[graphicx]{realboxes}
\usepackage{adjustbox}
\usepackage{tablefootnote}
\usepackage{tabularx}
\usepackage{multirow, makecell}
\usepackage{array}
\usepackage{diagbox}
\usepackage{enumerate}
\usepackage{enumitem}
\usepackage[caption=false,font=footnotesize]{subfig}
\usepackage{url}
\usepackage{textcomp}

\usepackage{algpseudocode}
\usepackage{color}
\usepackage[ruled,vlined]{algorithm2e}

\DeclareMathOperator*{\argmax}{argmax}

\def\BibTeX{{\rm B\kern-.05em{\sc i\kern-.025em b}\kern-.08em
    T\kern-.1667em\lower.7ex\hbox{E}\kern-.125emX}}

\begin{document}

\title{Sample Dominance Aware Framework via Non-Parametric Estimation for Spontaneous Brain-Computer Interface\\

\thanks{This work was supported by the Institute of Information \& Communications Technology Planning \& Evaluation (IITP) grant, funded by the Korea government (MSIT) (No. 2019-0-00079, Artificial Intelligence Graduate School Program (Korea University)).}
}

\author{\IEEEauthorblockN{Byeong-Hoo Lee}
\IEEEauthorblockA{\textit{Dept. of Brain and Cognitive Engineering} \\
\textit{Korea University}\\
Seoul, Republic of Korea \\
bh\_lee@korea.ac.kr}
\and
\IEEEauthorblockN{Byoung-Hee Kwon}
\IEEEauthorblockA{\textit{Dept. of Brain and Cognitive Engineering} \\
\textit{Korea University}\\
Seoul, Republic of Korea \\
bh\_kwo@korea.ac.kr}
\and
\IEEEauthorblockN{Seong-Whan Lee}
\IEEEauthorblockA{\textit{Dept. of Artificial Intelligence} \\
\textit{Korea University}\\
Seoul, Republic of Korea \\
sw.lee@korea.ac.kr}
}
\maketitle

\begin{abstract}
Deep learning has shown promise in decoding brain signals, such as electroencephalogram (EEG), in the field of brain-computer interfaces (BCIs). However, the non-stationary characteristics of EEG signals pose challenges for training neural networks to acquire appropriate knowledge. Inconsistent EEG signals resulting from these non-stationary characteristics can lead to poor performance. Therefore, it is crucial to investigate and address sample inconsistency to ensure robust performance in spontaneous BCIs. In this study, we introduce the concept of sample dominance as a measure of EEG signal inconsistency and propose a method to modulate its effect on network training. We present a two-stage dominance score estimation technique that compensates for performance degradation caused by sample inconsistencies. Our proposed method utilizes non-parametric estimation to infer sample inconsistency and assigns each sample a dominance score. This score is then aggregated with the loss function during training to modulate the impact of sample inconsistency. Furthermore, we design a curriculum learning approach that gradually increases the influence of inconsistent signals during training to improve overall performance. We evaluate our proposed method using public spontaneous BCI dataset. The experimental results confirm that our findings highlight the importance of addressing sample dominance for achieving robust performance in spontaneous BCIs.
\end{abstract}


\IEEEpeerreviewmaketitle

\section{INTRODUCTION}

\IEEEPARstart{D}{eep} learning has shown promising performance in classifying intricate data such as medical images \cite{thung2018conversion}, acoustic signals \cite{jia2018transfer}, and brain signals \cite{kim2019subject, li2018hybrid, Tabar}. It has achieved significant performance improvements when provided with enough training samples. Therefore, numerous studies have investigated augmentation techniques to obtain additional training samples \cite{lashgari2020data}. However, if the training dataset includes outliers and label noise samples, model performance will be decreased although a sufficient number of training samples have been collected \cite{lee2019possible, kim2015abstract}. 

\begin{figure*}[!t]
  \centerline{\includegraphics[scale =0.8]{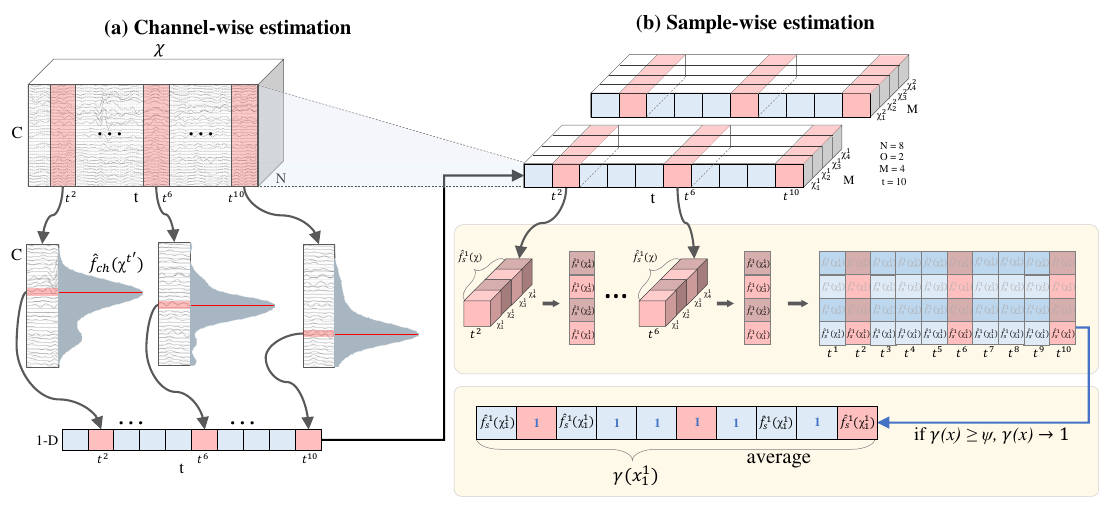}}
  \caption{Framework of the dominance score estimator. The encoder of the SAE obtains representations $\chi$ along the channel $C$. Thereafter using $\chi$, a two-stage dominance score estimation is performed based on the kernel density estimation. The channel-wise estimation selects a representative channel value that is assigned the highest probability by the estimated probability density function (PDF) for all encoded times $t$. An estimation procedure is performed through sample-wise estimation using the same class of samples $M$. The PDF is repeatedly estimated using a time point of all samples yielding $t$ numbers of probability $\hat{f_{s}}(\chi)$ per sample.}
\end{figure*}

In this study, we present the concept of sample dominance to address the aforementioned challenges of spontaneous BCIs. We assumed that the dominant samples are valid samples that are consistently generated, and the non-dominant samples are inconsistent samples, i.e. invalid samples. Furthermore, the dominant samples are clustered; meanwhile, the non-dominant samples are scattered, which interferes with the network training. Our approach involves inferring sample dominance to estimate the dominance score and utilize it to improve performance. Consequently, we proposed a non-parametric estimation-based framework that estimates the dominance score and aggregates it with the loss function. The proposed framework consists of a multilayer perceptron-based stacked autoencoder (SAE), dominance score estimator, and backbone network for decoding electroencephalogram (EEG) signals. The SAE produces representations of the input data through a bottleneck layer. The dominance score is estimated using a two-stage procedure based on non-parametric estimation, specifically, kernel density estimation (KDE) with a Gaussian kernel function \cite{KDE}. The objective is to map the distance between the samples to probability. Thus, according to the probability density function (PDF), clustered samples are assigned a high probability, whereas scattered samples are assigned a low probability. Valid and invalid time points are selected based on the confidence threshold. The probability of valid time points becomes 1, which indicates that they are fully considered for training. The average of all PDFs was considered as the dominance score per sample. Then the score was aggregated into a loss function to modulate the training loss. 

The main contributions of this study are as follows: \textit{i}) The concept of sample dominance was introduced to investigate sample inconsistency and its effect on performance was confirmed using public spontaneous BCI datasets. \textit{ii}) We then demonstrated that the dominance score improved performance by using curriculum learning, regardless of the type of spontaneous BCI. \textit{iii}) The proposed method was not only compatible with existing methods but also improved their performance.

\section{OVERALL FRAMEWORK}
\subsection{Sample Inconsistency in Spontaneous BCIs}
Spontaneous BCIs require users to consistently perform imagery tasks to minimize the factors that affect EEG signals because data collection depends on EEG signal recording technologies and also the ability of the user to induce EEG signals. Moreover, EEG signals have oscillatory waveforms, which makes it difficult to interpret and decode information. Additionally, the concentration and fatigue of users affect the generation of EEG signals \cite{suk2014predicting, lee2020neural, HSCNN}. Thus, it is difficult to guarantee that meaningful EEG signals that fit the label have been generated, despite users feeling that they have imagined correctly. Hence, the dataset contained noisy labels and outlier samples. Using these samples creates interference in network training, resulting in performance degradation. In this study, we considered dominant samples as valid; meanwhile, non-dominant samples were considered invalid, based on the unknown ground truth of the EEG signals.

\subsection{Self-Supervised Learning for Representation}
EEG signals are complicate signals, containing brain activity and background noise. Thus, these type of noise should be removed through the preprocessing prior to feature extraction for performance improvement. However, as we mentioned above, distinguishing noise and EEG signals is challenging work. Due to the uninterpretable characteristics of the EEG signals, it is difficult to confirm that the quantitative methods defined by a human is valid. Therefore, we conjecture that self-supervised learning is suitable for obtaining representations of EEG signals in terms of model training rather than supervised learning \cite{hendrycks2019using}. Representations are considered that they contain relevant features and reduce the computational costs of decoding procedure. To this end, we designed a multilayer perceptron-based SAE that consisted of two hidden layers with a ReLU \cite{mane2021fbcnet} activation function to obtain the representations. From the training dataset, the encoder $f_E$ learns mapping $X \rightarrow \chi$ by reducing the time dimension of the input data $X \in \mathbb{R}^{C\times T}$ $= [x_1, x_2, ..., x_N]$, where $C$, $N$ and $T$ denote the number of channels, samples and time points respectively. The $f_E$ firstly divides channels of $x$ into individuals and extract representations channel by channel. Then it produces $\chi$ by concatenating each representation on the channel axis. Design choices of the SAE is described in Table I. Therefore, the EEG representation $\chi \in \mathbb{R}^{C\times t}$ $= [\chi_1, \chi_2, ..., \chi_N]$ is obtained from $f_E$, wherein $t$ denotes the encoded time point \cite{lee2019towards}. In this study, {$t$ is one-quarter of $T$ according to the SAE. $\chi$ is used only for dominance score estimation and $x$ is fed into backbone network for training. The SAE is trained to reduce the difference between the original data and decoder output via the loss function, which is defined as

\begin{equation}
   \mathcal{L}_{SAE} = \frac{1}{N}\sum_{i=1}^{N} \left (Y_{i} - x_{i} \right )^{2},
\end{equation}
where $Y$ denotes the concatenated output of the decoder. As training progresses, the $f_E$ represents the input data at low dimensions through the bottleneck layer. The dominance score estimation was conducted using $\chi$.

\begin{figure}[!t]
  \centerline{\includegraphics[scale =0.9]{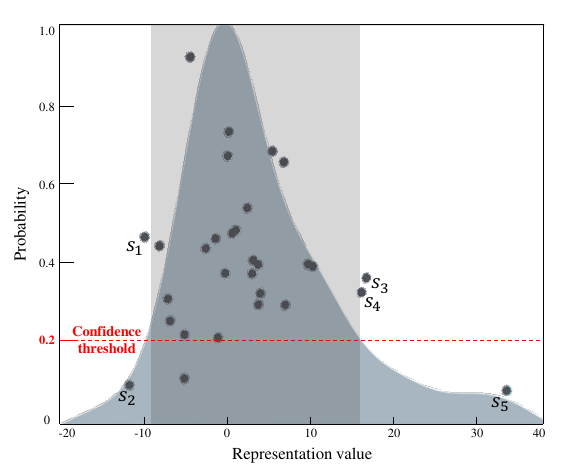}}
  \caption{Illustration of score estimation. For better visualization, we used two-dimensional representations. Most samples are assigned a value of 1, but only five samples ($s_{1}$ to $s_{5}$) are assigned a probability.}
\end{figure}

\subsection{Dominance Score Estimation}
The overall framework of dominance score estimation is shown in Fig. 1. Considering the non-stationary characteristics of EEG signals, the probability-based approach is advantageous for dominance score estimation. $X$ is distributed using an unknown PDF $f(X)$. The objective of this step is to obtain ${f}(X)$ by estimating the appropriate $\hat{f}(\chi)$ for dominance score estimation. The example of estimation is illustrated in Fig. 2.

\subsubsection{Channel-wise estimation}
Depending on the paradigm, EEG signals are strongly induced in specific brain areas (e.g., the sensorimotor cortex in motor imagery). Channel values are EEG signals recorded from the brain area wherein the channels are located. When users perform a paradigm, EEG signals are intensively induced from specific brain areas, and these areas change over time \cite{hohne2014motor, lee2019comparative}. Therefore, the channel values containing intensive EEG signals are clustered at a specific $t$. At this stage, the score estimator estimates the channel-wise PDF $\hat{f_{ch}}(\chi)$ to select a representative channel value for all $t$ values as depicted in the bottom part of Fig. 1(a). For instance, at an arbitrary time point $t'$, $\chi$ can be expressed as $\chi^{t'} = [\chi^{t'}_{1}, \chi^{t'}_{2}, \chi^{t'}_{3}, ... \ , \chi^{t'}_{C}]$. $\hat{f_{ch}}(\chi^{t'})$ is estimated based on every component of $\chi^{t'}$ which is defined as}

\begin{equation}
    \hat{f_{ch}}(\chi^{t'}) = \frac{1}{Ch} \sum_{i=1}^{C} K \left(\chi^{t'} - \chi^{t'}_{i} \over h \right ),
\end{equation}
where $K$ and $h$ denote the kernel function and the smoothing parameter, respectively. The kernel function selected for this study was the Gaussian kernel, which is defined as

\begin{equation}
    K(x) = \frac{1}{\sqrt{2\pi}} \exp \left (-\frac{1}{2} x^2 \right ),
\end{equation}
Therefore, (3) is applied to (2), and $\hat{f_{ch}}(\chi^{t'})$ is represented as a score estimator that repeats the procedure for all values of $t$ (center part of Fig. 1(a) as follows).
\begin{equation}
    \hat{f_{ch}}(\chi^{t'}) = \frac{1}{Ch} \sum_{i=1}^{C} \frac{1}{\sqrt{2\pi}} \exp \left (-\frac{\vert\vert \chi^{t'} - \chi^{t'}_{i} \vert\vert}{2h^2} \right )
\end{equation}

The $\chi^{t'}$ is now the value that derives the highest channel-wise PDF, determined as follows:  

\begin{equation}
    \chi^{t'} = \argmax_C \hat{f_{ch}}(\chi^{t'})
\end{equation}

Equation (5) is repeated for all $t$ thereby yielding $\chi \in \mathbb{R}^{1\times t}$. The goal of this stage is to obtain a single representative channel value to calculate the dominance score. However, multiple $\chi^{t'}$ can be selected according to (5). Therefore, the average of $\chi^{t'}$ is calculated as a representative value.

\subsubsection{Sample-wise estimation}
Sample-wise estimation is conducted using $\chi \in \mathbb{R}^{1\times t}$ that is estimated previous stage. The $\hat{f_s}$ is estimated by considering the values among the same class samples while the previous stage uses a single sample to estimate $\hat{f_s}$ considering channel values. To this end, $\chi$ is concatenated by class yielding $\chi^{o'} \in \mathbb{R}^{M\times t}$ where $M$ is the number of samples in class $o'$. For arbitrary class $o'$, $\chi^{o'}$ is expressed as $\chi^{o'} = [\chi^{o'}_1, \chi^{o'}_2, \chi^{o'}_3, ... , \chi^{o'}_M]$. The score estimator calculates the sample-wise PDF $\hat{f_{s}}(\chi^{o'})$ at a fixed time $t$ of $\chi^{o'}$. The procedure is repeated for all $t$; therefore, $\chi^{o'}$ is given $t$ numbers of PDF yielding $\hat{f_{s}}(\chi^{o'}) = [\hat{f_{s}}(\chi^{o'}_1), \hat{f_{s}}(\chi^{o'}_2), \hat{f_{s}}(\chi^{o'}_3), ... \ ,\hat{f_{s}}(\chi^{o'}_t)]$. In other words, every component of $\chi^{o'}$ is assigned $t$ numbers of $\hat{f_{s}}(\chi^{o'})$.  Similar to channel-wise estimation, $\hat{f_s}(\chi^{o'})$ is defined as

\begin{equation}
    \\hat{f_s^{o'}}(\chi) = \frac{1}{Mh} \sum_{i=1}^{M} K\left (\chi^{o'} - \chi^{o'}_{i} \over h \right ),
\end{equation}
where $O$ denotes the number of classes. The Gaussian kernel function (3) is applied to (6), and this formulation can be expressed as

\begin{equation}
   \hat{f_s}(\chi^{o'}) = \frac{1}{Mh} \sum_{i=1}^{M} \frac{1}{\sqrt{2\pi}} \exp \left (-\frac{\vert\vert \chi^{o'} - \chi^{o'}_i \vert\vert}{2h^2} \right ),
\end{equation}

As we mentioned above, each $\chi$ is assigned $t$ numbers of $\hat{f_s}(\chi)$. Dominance score $\gamma(x)$ is calculated based on the $\hat{f_s}(\chi)$. The objective of the proposed method is to modulate the influence on model training by assigning a single dominant score for each $x$. To this end, average of $\hat{f_s}(\chi)$ is assigned to $x$. This can be expressed as

\begin{equation}
  \gamma(x) = \frac{1}{t}\sum_{i=1}^{t} \hat{f_s}(\chi_t), \ (0 \leq \gamma(x) \leq 1),
\end{equation}
here, $\chi$ is a representation of corresponding $x$. Therefore, the $\gamma(x)$ is calculated based on $\chi$ that is assigned to the corresponding $x$.  Additionally, confidence threshold $\psi$ is introduced to reduce computation cost. The $\psi$ is a hyperparameter that modulates influence on $\gamma(x)$ on model training. For all $\gamma(x)$ which is included in $\psi$ are assigned 1 instead of its $\gamma(x)$ to indicate that $x$ is dominant, as shown in Fig. 2. Otherwise, $\gamma(x)$ is the corresponding $\hat{f_s}(\chi)$. Accordingly, most $x$ belong to $\psi$ and are assigned a value of 1, as shown in Fig. 2. However, five samples ($s_1$ - $s_5$) are given $\gamma(x)$ at this arbitrary time point. This can be expressed as

\begin{equation}
   \gamma(x) = 
    \begin{cases}
        1 & \text{if $\gamma(x) \geq \psi$} \\
        \gamma(x) & \text{otherwise}
    \end{cases}
\end{equation}
Consequently, each training sample is assigned its averaged $\gamma(x)$, which is designed to reduce the loss of non-dominant training samples. To this end, $\gamma(x)$ is multiplied by a loss function to reduce the training effect.

However, learning non-dominant samples can improve performance somehow; hence, the test dataset also contains non-dominant samples. Therefore, we designed a form of curriculum learning that gradually increases $\gamma(x)$ up to 1 to avoid a drastic change in loss, defined as

\begin{equation}
\gamma(x)=\begin{cases} \gamma(x) &       \text{$epoch < T_{1}$} \\
\gamma(x) + \frac{1- \gamma(x)}{T_{2}-T_{1}}  & \text{$T_{1}\leq epoch \leq T_{2}$}\\
1 &       \text{$T_{2} < epoch$}
\end{cases}
\end{equation}
where $T_{1}$ and $T_{2}$ are the start and end points of the score-increasing period during training epochs. In Fig. 2, only non-dominant samples ($s_1$ - $s_5$) are applied to (13). Therefore, non-dominant samples have a lower influence on training in the preliminary stages, and this influence gradually increases to enhance backbone network robustness. After $T_2$, the non-dominant samples are treated in the same manner as the dominant samples. The overall procedure for dominance score estimation is summarized in Algorithm 1. 

$\gamma(x)$ was estimated to reduce the effect of non-dominant samples on training. To this end, we designed a loss function such that $\gamma(x)$ is multiplied by the loss, which is defined as:

\begin{equation}
  \mathcal{L} = -\frac{1}{N} \sum_{j=1}^{N} \sum_{i=1}^{O}\gamma(x_j) y_{i} log(p_{i}),
\end{equation}
where $y$ and $p$ denote the true label and model prediction, respectively. According to (14), the loss function produces a loss, which is reduced by $\gamma(x)$. Thus, the training samples, which were considered non-dominant, produce smaller losses. Hence (13) gradually increases $\gamma(x)$ to 1, and the loss function remains the same as the standard cross-entropy loss after $T_{2}$.

\begin{table}[t!]
\centering
\renewcommand{\arraystretch}{1.1}
{
\caption{Hyperparameters for the experiments. $h$ was calculated using Equation (10). The size of $\chi$ was determined using the SAE and imagination period of datasets.}
{\normalsize
\begin{tabular}{ccccccc}
\cline{1-7}
\multicolumn{1}{c}{Datasets} & K        & $h$  & $t$ size & $\psi$ & $T_1$ & $T_2$\\\hline
\multicolumn{1}{l}{Dataset-I}        & Gaussian  & 3  & 250                 & 90   & 50 & 150 \\ \hline
\end{tabular}}}
\end{table}

\section{RESULTS AND DISCUSSION}
\subsection{Experimental protocols}
To evaluate the proposed method, we conducted experiments using public spontaneous BCI dataset: BCI Competition IV dataset-2a \cite{competition}. The sampling rate was 250 Hz thus, $t$ was 250-time points. The evaluation was conducted in a subject-dependent manner without advanced filtering methods that would affect the performance to obtain a standard performance \cite{bang2021spatio}. Several methods \cite{eegnet, deepconvnet, ICASSP, MCNN, siddhad2022efficacy} were selected for the backbone network. Both the deep and shallow ConvNets were selected from \cite{deepconvnet}. Additionally, we applied the data cropping method introduced in \cite{deepconvnet} using a sliding time window with a stride of 100 ms and cross-validation. Because the average of all crops was used as the final prediction, a decision was made for each sample. We set 200 and 500 training epochs for the backbone network and SAE training, respectively. The evaluation was conducted using the weights of backbone networks showed the lowest validation loss after 180 epochs and the weights of the SAE were obtained at 500 epochs. The AdamW optimizer \cite{AdamW} with a learning rate of 0.001 and a weight decay of 0.01 were used. The average classification accuracy of all folds was reported as the performance of the subjects. The configurations of the hyperparameters and the evaluation results are listed in Tables I and II, respectively. It is noted that Table II lists the subject-averaged accuracy. $h$ was rounded off to reduce calculation complexity. This experiment was conducted on a system comprising an Intel Core i7 9700 K CPU running at 3.60 GHz, 32 GB of DDR4 RAM, two NVIDIA TITAN V GPUs (1200 MHz for each), and Python version 3.7 with PyTorch version 1.6.

\begin{table}[t!]
\centering
\renewcommand{\arraystretch}{1.1}
\caption{Classification accuracy with/without a cropping method. Reported results are subject-averaged accuracy (\%) and standard deviation. ``()'' denotes standard deviation. Superscripts $^1$ and $^2$ denote shallow and deep Convnet \cite{deepconvnet}. The highest accuracy and the lowest standard deviation are denoted in bold.}
{\small
\begin{tabularx}{\columnwidth}{cccc} \hline
Method                                & \multicolumn{3}{c}{Accuracy}                           \\ \hline
\multirow{4}{*}{Lawhern \textit{et al.} \cite{eegnet}}   & \multirow{2}{*}{w/o crop.}  & Baseline  & 66.27 (10.88) \\
                                      &                             & with ours & 70.61 (12.00) \\ \cline{2-4}
                                      & \multirow{2}{*}{with crop.} & Baseline  & 71.13 (13.18) \\ 
                                      &                             & with ours & 73.25 (12.96) \\\hline
\multirow{4}{*}{Schirrmeister \textit{et al.}$^1$ \cite{deepconvnet}} & \multirow{2}{*}{w/o crop.}  & Baseline  & 66.03 (9.43)  \\
                                      &                             & with ours & 71.24 (10.04) \\\cline{2-4}
                                      & \multirow{2}{*}{with crop.} & Baseline  & 70.58 (17.87) \\
                                      &                             & with ours & 73.03 (17.92) \\\hline
\multirow{4}{*}{Schirrmeister \textit{et al.}$^2$ \cite{deepconvnet}} & \multirow{2}{*}{w/o crop.}  & Baseline  & 63.02 (15.12) \\
                                      &                             & with ours & 67.72 (13.80) \\\cline{2-4}
                                      & \multirow{2}{*}{with crop.} & Baseline  & 66.80 (11.63) \\
                                      &                             & with ours & 70.65 (13.50) \\\hline
\multirow{4}{*}{Lee \textit{et al.} \cite{ICASSP}}          & \multirow{2}{*}{w/o crop.}  & Baseline  & 65.93 (7.82)  \\
                                      &                             & with ours & 70.54 (8.76)  \\\cline{2-4}
                                      & \multirow{2}{*}{with crop.} & Baseline  & 68.31 (11.16) \\
                                      &                             & with ours & 71.98 (12.03) \\\hline
\multirow{4}{*}{Amin \textit{et al.} \cite{MCNN}}          & \multirow{2}{*}{w/o crop.}  & Baseline  & 66.21 (7.26)  \\
                                      &                             & with ours & 71.59 (5.15)  \\\cline{2-4}
                                      & \multirow{2}{*}{with crop.} & Baseline  & 71.08 (10.10) \\
                                      &                             & with ours & 74.33 (10.11) \\\hline
\multirow{4}{*}{Siddhad \textit{et al.} \cite{siddhad2022efficacy}}       & \multirow{2}{*}{w/o crop.}  & Baseline  & 66.21 (7,26)  \\
                                      &                             & with ours & 74.33 (10.11) \\\cline{2-4}
                                      & \multirow{2}{*}{with crop.} & Baseline  & 72.16 (9.43)  \\
                                      &                             & with ours & 75.69 (9.79)  \\\hline
\end{tabularx}}
\end{table}

\subsection{Performance without cropping}
According to Table II, $\gamma(x)$ improved the performance compared to the baseline. The proposed method achieved performance improvements of up to 5\% across all the backbone networks. Using the proposed method, the backbone network in Siddhad \textit{et al.} \cite{siddhad2022efficacy} obtained an accuracy of approximately 71\%, which was the highest. The one in Amin \textit{et al.} \cite{MCNN} showed the highest performance improvement (5.38\%), which is approximately only 1\% higher than that of the network in Lawhern \textit{et al.} \cite{eegnet}, which exhibited the lowest performance improvement. Thus, the proposed method consistently led to performance improvements across all backbone networks. 

\subsection{Performance with cropping}
The experimental results confirmed that the data cropping method improved the performance of the backbone networks. The network in Siddhad \textit{et al.} \cite{siddhad2022efficacy} obtained the highest accuracy (72.16\%) using only the cropping method. The data cropping method allows the backbone network to consider a single training sample multiple times. The final output of the backbone network was the average output of each crop. Using both $\gamma(x)$ and the data cropping method yielded the best performance, with performance improvements of up to 8.8\% compared with the baseline performance without the cropping method. Particularly, the backbone network in Siddhad \textit{et al.} \cite{siddhad2022efficacy} achieved the highest performance (75.69\%) and improvement (8.8\%). However, compared with the baseline using the cropping method, the average performance improvement using the proposed method was 3\%, and the shallow network in Schirrmeister \textit{et al.} \cite{deepconvnet} showed the highest performance improvement (3.85\%). 

\section{CONCLUSION}
In this study, we focused on estimating the dominance score by inferring the sample dominance for performance improvement. We proposed a two-stage dominance score estimation to modulate the effects of non-dominant samples during training. Based on the probability density function and its confidence threshold, non-dominant samples (noisy label or outlier samples) were assigned a dominance score. The score was aggregated with the loss function to decrease the loss. This reduced the influence of non-dominant samples on training. However, because they help in improving network performance, the test dataset also contained non-dominant samples. Through an experiment, we demonstrated that the proposed method can improve network performance despite spontaneous paradigms. However, spontaneous BCIs still have challenging issues, as we discussed. Therefore, future works will backtrack the origin of EEG and investigate the differences between neuroscientific facts and electrode-based EEG recordings.
\bibliographystyle{IEEEtran}
\bibliography{MANUSCRIPT}
\end{document}